%% file: root.tex
\documentclass[letterpaper, 10 pt, conference]{ieeeconf}
\usepackage{times}

\IEEEoverridecommandlockouts                              %
                                                          
\usepackage[bookmarks=true]{hyperref}

\usepackage{siunitx}
\usepackage{amsmath}
\usepackage{amssymb}
\usepackage{amsfonts}
\DeclareMathAlphabet{\mathcal}{OMS}{cmsy}{m}{n}

\usepackage{nicefrac}
\usepackage{soul}
\usepackage{graphics} %
\usepackage{epsfig} %
\usepackage{mathptmx} %
\usepackage{times} %
\usepackage[colorinlistoftodos,prependcaption,textsize=tiny,disable]{todonotes}
\usepackage{hyperref}
\usepackage[export]{adjustbox}

\usepackage{float}
\usepackage{xcolor}%
\usepackage{colortbl}
\usepackage{tikz,array,collcell}
\usepackage{pifont}
\usepackage{ragged2e}
\usepackage{subcaption}
\usepackage{booktabs}
\usepackage{tabularx}
\usepackage[font=small]{caption}

\usepackage{threeparttable}

\usepackage[strict]{changepage}

\usepackage{algorithm}
\usepackage{algpseudocode}

\makeatletter
\patchcmd{\@makecaption}
  {\scshape}
  {}
  {}
  {}
\makeatletter
\patchcmd{\@makecaption}
  {\\}
  {.\ }
  {}
  {}
\makeatother

\usepackage{xspace}

\input{notation.tex}

\title{\LARGE \bf
Inferring Articulated Rigid Body Dynamics \\from RGBD Video
}

\author{
Eric Heiden${}^{1}$, Ziang Liu${}^{2}$, Vibhav Vineet${}^{3}$, Erwin Coumans${}^{1}$, Gaurav S. Sukhatme${}^{4}$%
\thanks{$^{1}$NVIDIA, Santa Clara, USA
        {\tt\small eheiden@nvidia.com}}%
\thanks{$^{2}$Stanford University, Stanford, USA}%
\thanks{$^{3}$Microsoft Research, Redmond, USA}%
\thanks{$^{4}$Google Research, Mountain View, USA}%
\thanks{$^{4}$G.S. Sukhatme holds concurrent appointments as a Professor at the University of Southern California (USC) and as an Amazon Scholar. This paper describes work performed at USC and is not associated with Amazon.}%
\thanks{This work was supported by a Google PhD Fellowship.}
}

\usepackage{xspace}

\begin{document}

\maketitle
\thispagestyle{empty}
\pagestyle{empty}

\newcommand{\traj}[0]{\zeta}

\begin{abstract}
Being able to reproduce physical phenomena ranging from light interaction to contact mechanics, simulators are becoming increasingly useful in more and more application domains where real-world interaction or labeled data are difficult to obtain. Despite recent progress, significant human effort is needed to configure simulators to accurately reproduce real-world behavior.
We introduce a pipeline that combines inverse rendering with differentiable simulation to create digital twins of real-world articulated mechanisms from depth or RGB videos. Our approach automatically discovers joint types and estimates their kinematic parameters, while the dynamic properties of the overall mechanism are tuned to attain physically accurate simulations.
Control policies optimized in our derived simulation transfer successfully back to the original system, as we demonstrate on a simulated system. Further, our approach accurately reconstructs the kinematic tree of an articulated mechanism being manipulated by a robot, and highly nonlinear dynamics of a real-world coupled pendulum mechanism. \\
Website: {\footnotesize\url{https://eric-heiden.github.io/video2sim}}
\end{abstract}

\IEEEpeerreviewmaketitle

\section{Introduction}
\label{sec:intro}

From occupancy grids to modern SLAM pipelines, scene representations in robotics are becoming increasingly capable of informing complex behaviors. Approaches such as Kimera~\cite{rosinol2021kimera} are equipping robots with a world model that allows them to intelligently reason about spatiotemporal and semantic concepts of the world around them. While still centered around metric world representations, the interest in dynamics-aware representations is increasing.

Simulators, especially those that incorporate dynamics models, are world representations that can potentially reproduce a vast range of behavior in great detail. Provided these models are calibrated correctly, they can generalize exceptionally well compared to most purely data-driven models. They are an indispensable tool in the design of machines where the cost of prototyping hardware makes iterating in the real world prohibitively expensive. Robot control pipelines are often trained and developed in simulation due to the orders of magnitudes of speed-ups achievable by simulating many interaction scenarios in parallel faster than real time without causing damage in the early phases of training.

Nonetheless, it remains a challenge to leverage such tools for real-world robotic tasks. Not only is there a sim2real gap due to inherent model incompleteness, but it is often difficult to find the correct simulation settings that yield the most accurate results.
Despite recent advances in bridging the sim2real gap (see~\cite{hofer2020sim2real} for an overview), deriving a Unified Robot Description Format (URDF) file or analogous scene specifications poses a challenge when a real-world system needs to be simulated accurately.

We tackle the problem of automatically finding the correct simulation description for real-world articulated mechanisms. Given a depth or RGB video of an articulated mechanism undergoing motion, our pipeline determines the kinematic topology of the system, i.e. the types of joints connecting the rigid bodies and their kinematic properties, and the dynamical properties that explain the observed physical behavior. Relying on camera input our pipeline opens the avenue to future work integrating simulators more firmly into the representation stack that robots can use to reason about the physical world around them and make high-level decisions leveraging semantic information that the simulator encodes.

\begin{figure}
    \centering
    \includegraphics[width=0.75\columnwidth]{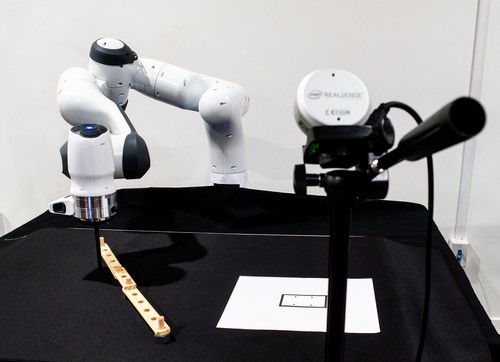}\vspace*{0.5em}
    \caption{\textbf{Experimental setup.} A Panda robot arm pushing an articulated system consisting of wooden toy construction parts via a cylindrical tip. The motion is recorded by an Intel RealSense L515 LiDAR sensor.}
    \label{fig:craftsman_setup}
\end{figure}

\begin{table*}[t!]
    \centering
    \resizebox{0.8\textwidth}{!}{
\begin{threeparttable}
    \begin{tabular}{lcccccccccc}
    \toprule
    & Graph NN~\cite{sanchez2018graph,sanchez2020learning} & gradSim~\cite{murthy2021gradsim} &  VRDP~\cite{ding2021vrdp} & Galileo~\cite{wu2015galileo} & ScrewNet~\cite{jain20screwnet} & Ours \\\midrule
        Infers joint topology & implicitly & & & & \checkmark & \checkmark\\
        Articulated physics & \checkmark & \checkmark & & & & \checkmark \\
        Differentiable physics & \checkmark & \checkmark & \checkmark & & & \checkmark\\
        Differentiable rendering & & \checkmark & & & & \checkmark \\
        Real-world ground-truth & \checkmark & & & \checkmark & \checkmark & \checkmark \\
        Pixel-based observations & & \checkmark & \checkmark & \checkmark & \checkmark & \checkmark \\
\bottomrule
    \end{tabular}
\end{threeparttable}
}\vspace*{0.5em}
    \caption{Comparison of selected inference and modeling approaches that reduce the gap between the real and simulated world.}
    \label{tab:related-works}
\end{table*}

By inferring the number of rigid objects and their relationship to each other without manually specifying such articulations upfront,  our work extends previous works that tackle system identification, i.e. the estimation of dynamical properties of a known model of the physical system. Our proposed pipeline consists of the following four steps:
\begin{enumerate}
    \item We identify the type and segmentation mask of rigid shapes in one frame of the video.
    \item We instantiate a differentiable rasterizer on a scene that consists of the previously identified shapes. Given the depth image sequence from the real system, we optimize the 3D poses of such rigid bodies via gradient-based optimization.
    \item We identify the types of joints connecting the rigid bodies based on their motion, and determine the kinematic properties of the mechanism.
    \item We infer the dynamical simulation parameters through a gradient-based Bayesian inference approach that combines differentiable rendering and simulation.
\end{enumerate}

Our experiments demonstrate the effectiveness of our proposed pipeline, as well as show the benefits of learning a physical simulator for control applications, and prediction of the behavior of real articulated systems.

\section{Related Work}
\label{sec:related}

In \autoref{tab:related-works} we provide a high-level overview of selected related works.

\subsection{Articulation inference}
\label{sec:rw-articulation}

Inferring the type and parameters of joints by observing moving bodies has been an important area in robotic perception, where affordances of objects need to be known for a robot to interact with them.
This has motivated interactive perception approaches, where action selection is informed by particle filtering~\cite{hausman2015articulation} or recursive Bayesian filtering (RBF)~\cite{eppner2018physics}. The latter approach builds on earlier work~\cite{martin2014articulation} that we adapt in our approach to find the parameters of revolute and prismatic joints given the motion of the rigid transforms of the objects in the scene (see \autoref{sec:joint-id}).

Articulation inference often hinges on accurate pose tracking of the rigid bodies, which is why many early works relied on fiducial markers to accurately track objects and subsequently infer articulations~\cite{sturm2011articulation, niekum2015online,liu2019articulation}. We do not require markers but do need to have accurate 3D meshes of the objects to be tracked in our approach.
Learning-based approaches, such as ScrewNet~\cite{jain20screwnet} and DUST-net~\cite{jain2021distributional} infer single articulations from depth images, whereas our approach recovers multiple articulations between an arbitrary number of rigid objects in the scene.
In \cite{mu2021asdf} and \cite{noguchi2021watch} signed distance fields are learned in tandem with the articulation of objects to infer kinematic 3D geometry, without considering the dynamics of the system.

\subsection{Learning simulators}
\label{sec:rw-learning}
Our approach instantiates a simulation from real-world observations, where the inference of articulations is the first step. Alternative approaches propose different pipelines to the problem of learning a simulator.
Entirely data-driven physics models often leverage graph neural networks to discover constraints between particles or bodies~(\cite{battaglia2016interaction, xu2019physics, sanchezgonzalez2020learning}). Physics-based machine learning approaches introduce a physics-informed inductive bias for learning models for dynamical systems from data~\cite{raissi2019physics,lutter2019delan,sutanto20rnea}.

Our work is closer to VRDP~\cite{ding2021vrdp}, a pipeline that leverages differentiable simulation to learn the parameters underlying rigid body dynamics. Contrary to our approach, VRDP does not consider articulated objects. Galileo~\cite{wu2015galileo} infers physical dynamics from video via Markov Chain Monte Carlo (MCMC) that samples a simulator as likelihood function, which, in contrast to our work, is not evaluated in image space but on low-dimensional pose tracking data and requires the full simulation with all objects to be manually specified.

\subsection{Differentiable simulation}
\label{sec:rw-diffsim}

Differentiable simulation has gained increasing attention, as it allows for the use of efficient gradient-based optimization algorithms to tune simulation parameters or control policies \cite{peres2018lcp, qiao2020scalable, geilinger2020add, murthy2021gradsim}, as well as LiDAR sensor models~\cite{heiden2020lidar}.

A particularly related prior work is gradSim~\cite{murthy2021gradsim} that, as in our approach, combines a differentiable physics engine with a differentiable renderer to solve inverse problems from high-dimensional pixel inputs. As in the other related works on differentiable simulators in this review, the simulator needs to have been set up manually before any optimization can begin. This entails that a URDF or other scene description format has to be provided that encodes the system topology and parameters. In this work, we introduce a method that automatically discovers the joint connections between rigid bodies from a mechanism observed through high-level inputs, such as depth or RGB camera images.

\section{Approach}
\label{sec:approach}

\begin{figure}
    \centering
    \includegraphics[width=0.8\columnwidth]{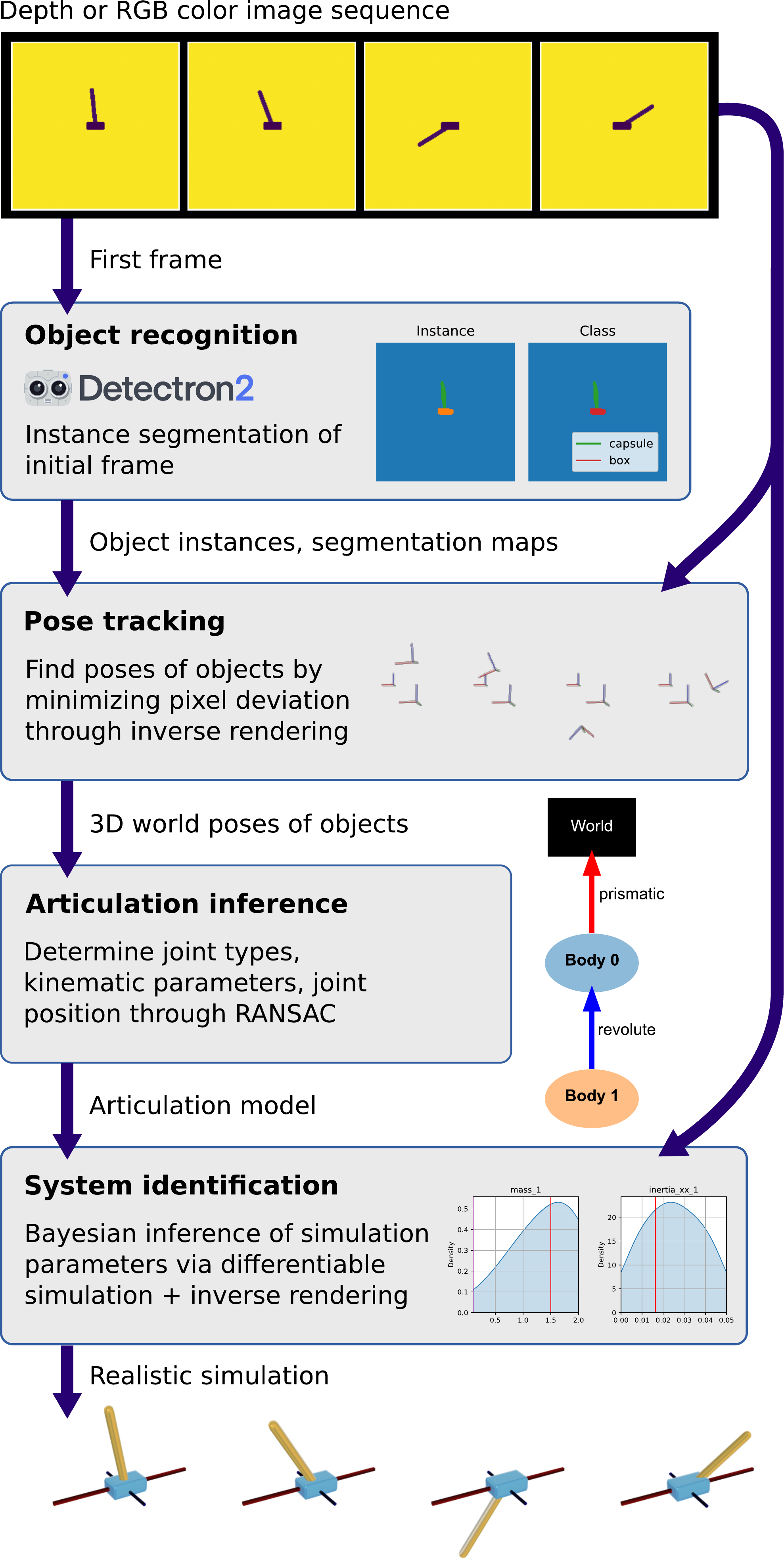}
    \caption{\textbf{Pipeline.} Our proposed simulation inference approach derives an articulated rigid body simulation from pixel inputs. The shown exemplary results generated by the phases in this diagram stem from the cartpole inference experiment from~\autoref{sec:exp-cartpole}.}
    \vspace{-0.05in}
    \label{fig:pipeline}
\end{figure}

In the following we describe the pipeline of our approach, which we summarize in~\autoref{fig:pipeline}.

\subsection{Geometry identification}
\label{sec:geom-id}

\begin{figure}
    \centering
    \includegraphics[width=0.7\columnwidth]{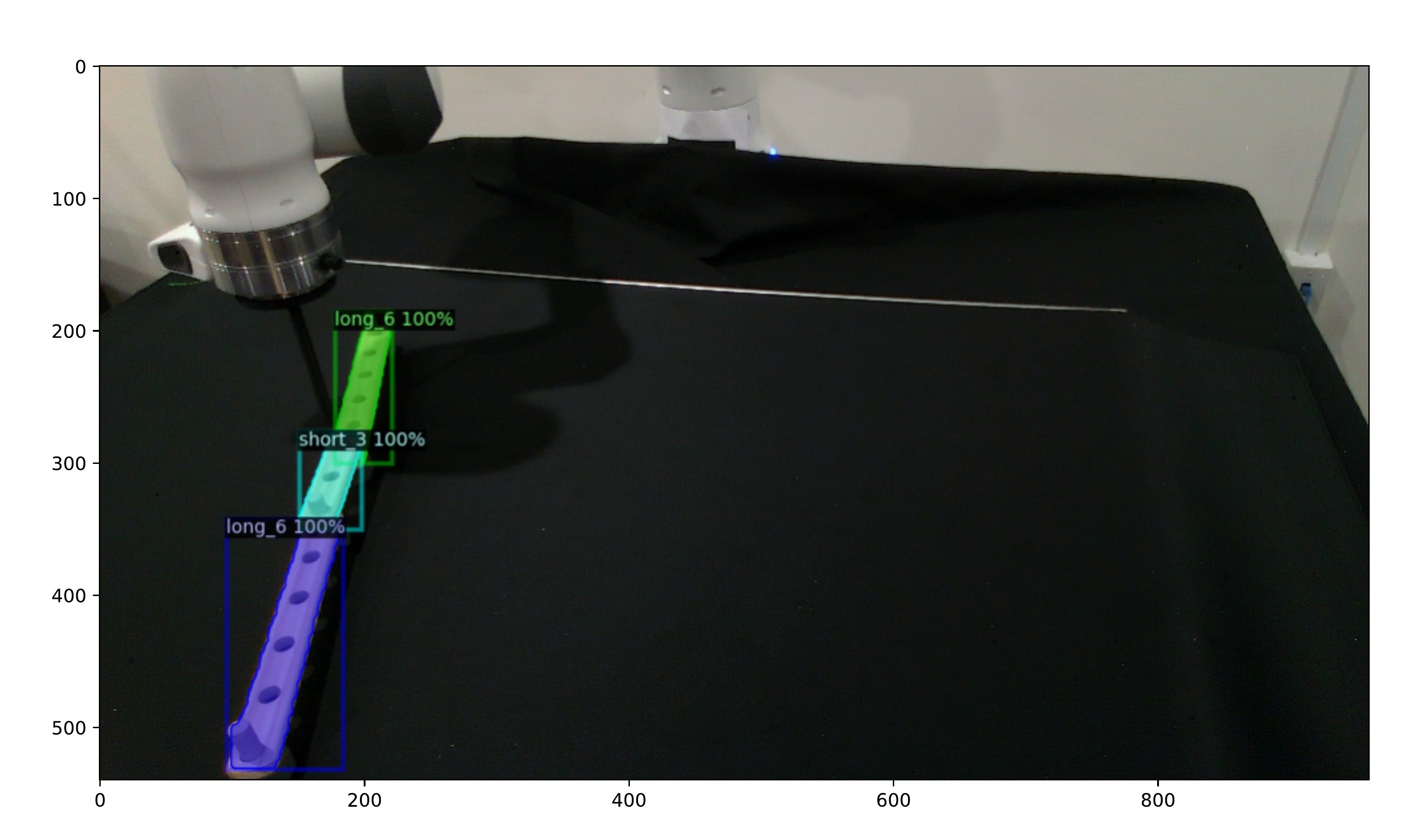}
    \caption{\textbf{Segmentation map prediction.} Detectron2 segmentation map predicted from an RGB image of a mechanism made of wooden parts from the Craftsman toolkit (see \autoref{sec:exp-craftsman}). The three pieces and their mesh types (either \texttt{long\_6} or \texttt{short\_3}) have been correctly identified.}
    \label{fig:craftsman-segmentation}
    \vspace{-0.05in}
\end{figure}

In the first phase, we use Detectron2~\cite{wu2019detectron2}, an object detection and instance segmentation network, to find the geometry instances and their segmentation masks in each frame of the input video. We found input images with three channels per pixel necessary to yield accurate results, so that we converted depth images to normal maps via finite differencing in pixel space. We generated two datasets in this work to train Detectron2: a synthetic set of normal maps of primitive shapes (capsules, boxes, spheres) of various sizes and in various pose configurations, and a dataset of camera images for our real-robot experiment in \autoref{sec:exp-craftsman}. An exemplar segmentation map predicted by Detectron2 on our Craftsman experiment is shown in~\autoref{fig:craftsman-segmentation}.

\subsection{Rigid body tracking}
\label{sec:rigid-tracking}

Given the segmentation map from the first step, we track the poses of the rigid bodies over time. We leverage nvdiffrast~\cite{laine2020nvdiffrast}, an efficient, GPU-driven differentiable rasterizer, to generate an observation $\observationvec$ (can be either a depth or RGB color image) given the 3D vertices of the object meshes $i\in{1,\dots,M}$ corresponding to the rigid bodies of interest in the scene. Such 3D vertices are transformed based on the world poses $T_0^i$ of the shapes we want to track. To solve the inverse rendering problem, i.e., finding object poses given pixel inputs, we minimize the L2 distance between the real image and the simulated rendering at each frame of the input video:
\begin{align*}
    \operatorname{minimize}_{[T_0^1, T_0^2, \dots, T_0^M]} \|f_{\text{rast}}([T_0^1, T_0^2, \dots, T_0^M]) - \observationvec\|^2,
\end{align*}
where $f_{\text{rast}}$ denotes the differentiable rasterizer function.

Note that our approach requires the background of the scene, i.e., the static bodies not part of the system, to be instantiated in the simulator if they influence the dynamics. In the case of a table-top manipulation task (e.g., the one from~\autoref{fig:craftsman_setup}), this may be the floor plane which is necessary for contact mechanics. Furthermore, the camera pose needs to be known upfront, and the 3D meshes of the objects in the mechanism have to be given.

\subsection{Joint identification and kinematic tracking}
\label{sec:joint-id}

We use the rigid pose trajectories to infer the types of joints that connect the rigid bodies, as well as the joint parameters (such as axis of rotation, pivot point, etc.). We follow a RANSAC approach~\cite{fischler1981ransac} that yields robust results despite noisy pose predictions. For each pair of rigid bodies, we set up one RANSAC estimator per joint type (model equations follow below): revolute, prismatic, and static (fixed) joint. We select the joint model with the least model error found after the RANSAC iterations. Given the guess for the joint type and its parameters, we compute the joint positions $q$ over time.

Given the sequence of world transforms $T_{0}^{i}$ of all rigid bodies $i$ in the scene, in the joint identification step we determine for each unique pair of rigid bodies $i$ and $j$ the relative transform $T_{i}^{j}[t]$ at each time step $t$. In the following, $T.r$ denotes the 3D axis-angle rotation vector, and $T.p$ the translation vector of a rigid pose $T$.

\paragraph{Revolute joint model}
Determine joint axis $\mathbf{s}$, pivot point $\mathbf{p}$ and joint angle $q$ from two consecutive transforms:
\begin{align*}
    \Delta r &= T_{i}^{j}[t\!+\!1].r - T_{i}^{j}[t].r~~~~~\Delta p = T_{i}^{j}[t\!+\!1].p - T_{i}^{j}[t].p\\
    \mathbf{s} &= \frac{\Delta r}{\| \Delta r \|}~~~~~~~\mathbf{p} = T_{i}^{j}[t].p + \frac{\Delta r \times \Delta p}{\| \Delta r \|^2}~~~~~~~q = \| \Delta r \|
\end{align*}

\paragraph{Prismatic joint model}
Determine joint axis $\mathbf{s}$ and joint position $q$ from two consecutive transforms:
\begin{align*}
    \Delta p &= T_{i}^{j}[t\!+\!1].p - T_{i}^{j}[t].p\\
    \mathbf{s} &= \frac{\Delta p}{\| \Delta p \|}~~~~~~~q = \mathbf{s} \cdot T_{i}^{j}[t\!+\!1].p
\end{align*}

\paragraph{Static joint model}
The static joint is parameterized by the relative transform $T_i^j$ between bodies $i$ and $j$.

\begin{algorithm}
\caption{Determine articulation of observed motion}
\label{alg:articulation-inference}
\begin{algorithmic}
\State \textbf{Input:} world transform sequence $T_{0}^{i}$ for each rigid body $i=1..n$
\State $C = \mathbf{0}_{n\times n}$
\For{body $i\in\{1..n\}$}
    \For{body $j\in\{i\!+\!1..n\}$}
        \State Calculate relative transform sequence $T_{i}^{j}$
        \State Determine joint parameters $\theta_{joint}$ for revolute (r), prismatic (p), static (s) joint type given $T_{i}^{j}$ via RANSAC, as well as their respective model errors $c_r, c_p, c_s$
        \If{RANSAC found at least one joint candidate}
            \State $C[i,j] = C[j,i] = \min \{c_r, c_p, c_s\}$
            \State Memorize $\theta^*_{joint}$ of candidate with lowest cost
        \Else
            \State $C[i,j] = C[j,i] = \infty$
        \EndIf
    \EndFor
\EndFor
\State Determine minimum spanning forest on $C$, retrieve root bodies $I_{root}$
\For{body $i\in I_{root}$}
    \State Construct kinematic tree $A$ rooted at body $i$, with parent-child connections from the corresponding minimum spanning tree and respective memorized joint parameters
    \State Find joint candidate $\theta_{joint}$ via RANSAC given $T_{0}^{i}$
    \If{RANSAC found at least one joint candidate}
        \State Attach $A$ to world via lowest-cost joint model $\theta^*_{joint}$
    \Else \Comment{floating-base case}
        \State Attach $A$ to world via free joint
    \EndIf
\EndFor
\State \textbf{return} world model consisting of articulations
\end{algorithmic}
\end{algorithm}

\begin{figure*}
    \centering
    \newcommand{\figheight}{3cm}
    \begin{subfigure}[b]{0.35\textwidth}
        \centering
        \hspace*{-1em}
        \includegraphics[height=\figheight,trim=0 0 4.5cm 0,clip]{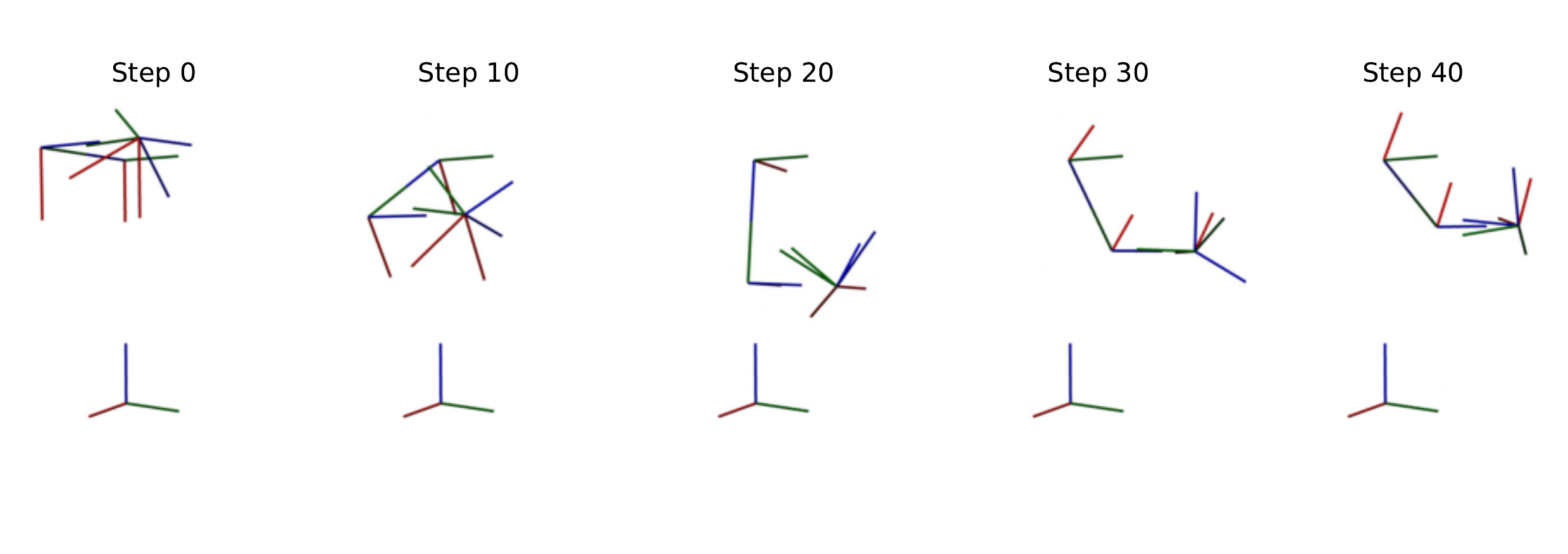}
        \caption{Body transforms}
        \label{fig:articulated-tree-tf-trajectory}
     \end{subfigure}
    \begin{subfigure}[b]{0.15\textwidth}
        \centering
        \hspace*{1em}
        \includegraphics[height=4cm]{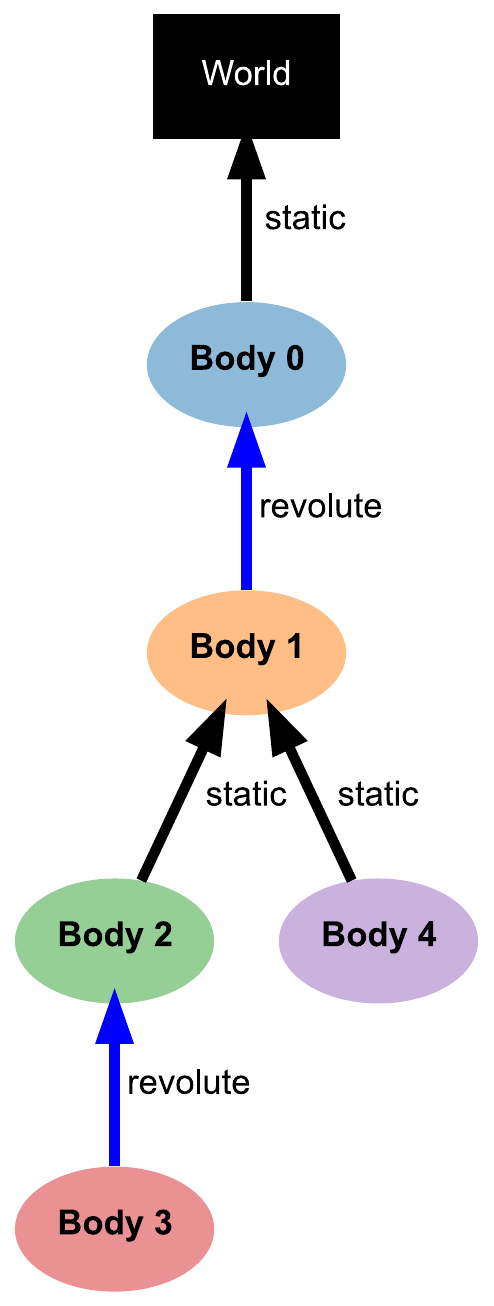}
        \caption{Inferred articulation}
        \label{fig:articulated-tree-scene-graph}
     \end{subfigure}
    \begin{subfigure}[b]{0.2\textwidth}
        \centering
        \includegraphics[height=\figheight]{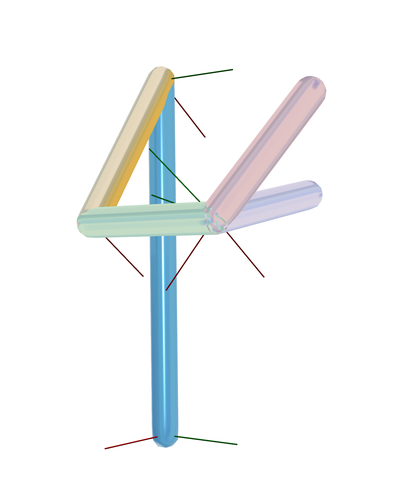}
        \caption{Reconstructed mechanism}
        \label{fig:articulated-tree-system}
     \end{subfigure}
    \begin{subfigure}[b]{0.25\textwidth}
        \centering
        \hspace*{-1.5em}
        \includegraphics[width=1.2\textwidth]{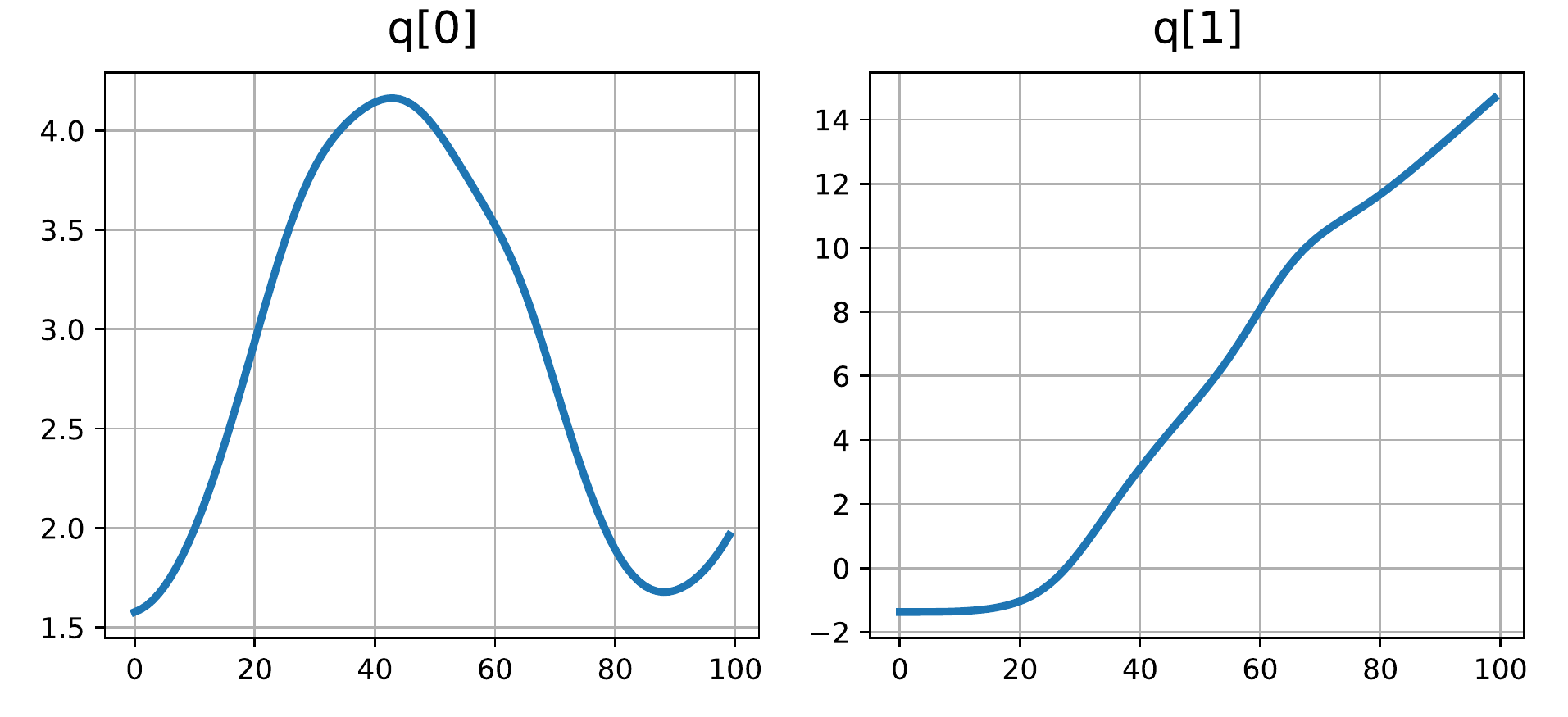}
        \caption{Joint positions}
        \label{fig:articulated-tree-qs}
     \end{subfigure}
    \caption{\textbf{Inference of articulations in a complicated mechanism consisting of static and revolute joints.}}
    \label{fig:articulated-tree}
\end{figure*}

Algorithm~\autoref{alg:articulation-inference} summarizes our inference approach to determine the articulations between rigid bodies given their observed motions. We first find the most likely joint types and corresponding joint parameters between unique pairs of rigid bodies via RANSAC for the three different joint models (revolute, prismatic and static). If no joint model could be found that matches the relative transform sequence between two rigid bodies, they are considered to be disconnected.

Having computed the cost matrix $C$ of the joint model errors from the previous RANSAC estimation, we find the minimum spanning forest via Prim's algorithm that we run on each component of the undirected graph described by the weighted adjacency matrix $C$. We select the root node $i$ from each minimum spanning tree as the top-level body in the kinematic tree to which all the rigid bodies within the same component are connected via the previously found joint models according to the hierarchy of the spanning tree. Note that since the joint models have been determined for a particular ordering of unique pairs of rigid bodies (to avoid duplicate computation), the direction the spanning tree imposes on the particular connection may require the joint model (e.g., in the case of revolute joints, the joint axis and pivot point) to be inverted.

Given the root node's sequence of world transforms $T_0^i$, we determine the most likely joint model again via RANSAC. If such a model has been found, the corresponding articulated mechanism is considered fixed-base and gets connected through this joint to the world. If no such joint model could be found, the articulation is floating-base and needs to be considered as such in the world model (either by adding degrees of freedom corresponding to a rigid-body motion, or via a flag that ensures the mechanism is simulated as a floating-base system).

\subsection{Simulation parameter estimation}
\label{sec:param-estimation}

Thus far, the pipeline was dedicated to the inference of the kinematic properties of the mechanism we observed. Having found the joint topology and approximate transforms between the bodies in the system, we will now consider the dynamics of the mechanism. This entails finding simulation parameters and an initial state vector that yield a motion, which, when rendered via our differentiable rasterizer, closely matches the given image sequence.

Leveraging Bayesian inference, we infer the dynamical and (optionally, for fine-tuning) kinematic parameters, such as joint axes and pivot points, of the mechanism.
We model the inference problem as a Hidden Markov Model (HMM) where the observation sequence $\trajectory=[\observationvec_1,\dots,\observationvec_T]$ of $T$ video frames is derived from latent states $\statevec_t$ ($t \in [1..T]$).
We assume the observation model is a deterministic function which is realized by the differentiable rasterization engine that turns a system state $\statevec_t$ into an observation image $\observationvec_t$ (either depth or RGB color).
The states are advanced through the dynamics model which we assume is fully dependent on the previous state and the simulation parameter vector $\params$. In our model, this transition function is assumed to be the differentiable simulator that implements the articulated rigid-body dynamics equations and contact models.
We use the Tiny Differentiable Simulator~\cite{heiden2021neuralsim} that implements end-to-end differentiable contact models and articulated rigid-body dynamics following Featherstone's formulation~\cite{featherstone2007rbda}.
For an articulated rigid body system, the parameters $\params$ may include the masses, inertial properties and geometrical properties of the bodies in the mechanism, as well as joint and contact friction coefficients.

Following Bayes' law, the posterior $p(\params|\trajectory)$ over simulation parameters $\params\in\mathbb{R}^\paramdim$ is calculated via $p(\params|\trajectoryset) \propto p(\trajectoryset|\params)p(\params)$.

We leverage the recently introduced Constrained Stein Variational Gradient Descent (CSVGD) algorithm~\cite{heiden2022pds} that adds constraints to the gradient-based, nonparametric Bayesian inference method SVGD.
Such constraint handling allows us to enforce parameter limits and optimize simulation parameters via \emph{multiple shooting}. This technique splits up the trajectory into shooting windows for which the start states need to be learned. Defect constraints are introduced that enforce continuity at the start and end states of adjacent shooting windows. Despite requiring extra variables to be optimized, multiple shooting significantly improves the convergence of gradient-based parameter inference approach when parameters need to be inferred from long time horizons.

\subsection{Limitations}
\label{sec:limitations}
We assume the possible geometric shapes of the rigid bodies of interest in the scene are known, so that they can be rendered. The sensor remains at a static pose, its intrinsics are known. The mechanism to be simulated must be in the viewport at least at the beginning of the video.
The tracking and inference currently does not run in real time.

\section{Experiments}
\label{sec:experiments}

We evaluate our approach on three mechanisms.

\subsection{Simulated Cartpole}
\label{sec:exp-cartpole}

\begin{figure}
    \centering
    \includegraphics[width=0.5\columnwidth]{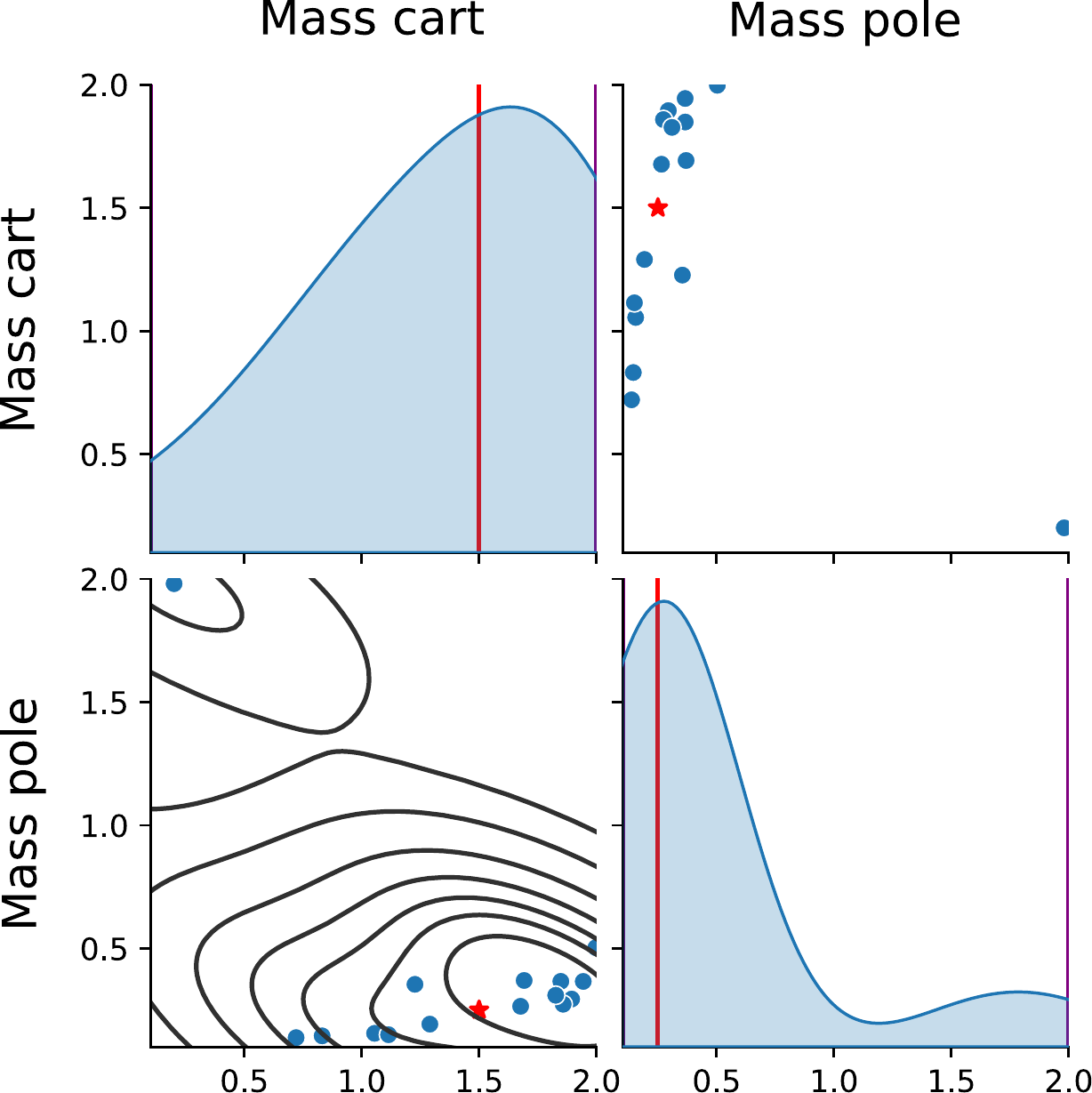}\vspace*{0.5em}
    \caption{\textbf{Parameter posterior distribution of the two link masses of the cartpole.} Inference from depth video via the multiple-shooting Bayesian inference approach CSVGD. The red lines and stars indicate the ground-truth parameters.}
    \label{fig:cartpole-mass-inference}
\end{figure}

In our first experiment we consider a simulated cartpole in the Bullet~\cite{coumans2013bullet} physics engine, which allows us to evaluate the accuracy of our method against the known ground-truth simulation parameters. As shown through the exemplary results in~\autoref{fig:pipeline}, we follow our pipeline to infer a realistic simulation given a sequence of 200 depth images. The cart and pole are correctly identified by Detectron2 as a box and capsule, which are then tracked via our inverse rendering approach. The subsequent articulation inference step find the expected joint topology where the pole is connected through a revolute joint with the cart, which in turn is connected by a prismatic joint to the world. Finally, we infer the simulation parameters related to the inertial properties of the two links (masses and inertia matrix diagonal entries). We visualize the posterior distribution found by CSVGD over the two link masses in~\autoref{fig:cartpole-mass-inference}, and compare the error over the inferred parameters compared to other methods in the top row of~\autoref{tab:cartpole-control}, which demonstrates that our approach is able to closely recover the true parameter settings (shown in red). In particular, the gradient-based estimators Adam and CSVGD achieve a lower normalized (w.r.t. parameter limits) parameter error within 2000 optimization steps than the CMA-ES~\cite{hansen2016cma} and MCMC method. We choose 16 particles both for CSVGD and CMA-ES. For MCMC, we also use 16 particles via the parallel, \emph{smooth-step} sampling scheme from the Emcee~\cite{foreman2013emcee} library. We select the particle with the highest likelihood for the comparison in~\autoref{tab:cartpole-control}.

\begin{table}[]
    \centering
    \resizebox{\columnwidth}{!}{
    \begin{tabular}{lccccc}
        \toprule
         & \bf Adam & \bf Ours & \bf CMA-ES & \bf MCMC & \bf GT \\\midrule
        \multicolumn{6}{l}{\bf Parameter error:}\\
        \bf NMAE & 0.299 & 0.161 & 0.369 & 0.328 & 0 \\ \midrule
        \multicolumn{6}{l}{\bf Average control reward (mean$\pm$std dev):}\\
        \bf Swing-up & $0.222\pm0.072$ & $0.235\pm0.059$ & $0.190\pm0.104$ & $0.229\pm0.071$ & $0.230\pm0.077$ \\
        \bf Balancing & $0.734\pm0.081$ & $0.742\pm0.081$ & $0.779\pm0.073$ & $0.744\pm0.128$ & $0.751\pm0.093$ \\
        \bottomrule
    \end{tabular}
    }\vspace*{0.5em}
    \caption{\textit{Top section:} normalized mean absolute error on the inferred parameters of the various methods compared against the true parameters. \textit{Bottom section:} average reward statistics obtained from 10 training runs of the MPPI controller evaluated on the Bullet cartpole system from \autoref{sec:cartpole-control} on the swing-up and balancing tasks.}
    \label{tab:cartpole-control}
\end{table}

\subsection{Rott's Chaotic Pendulum}
\label{sec:exp-rott}

In our next experiment, we aim to identify a real-world chaotic mechanism -- a coupled pendulum first analyzed by Nikolaus Rott~\cite{rott1970pendulum}. Rott's mechanism consists of two pendula with resonant frequencies at approximately two to one. One L-shaped pendulum is attached to a fixed pivot via a revolute joint, and a single body is attached to this L-shaped pendulum via another revolute joint. Given a video taken with an RGB camera of such a mechanism, we aim to reconstruct a digital twin in simulation.

In the first step of our pipeline, we identify the rigid objects in the video by converting the RGB images to binary images with an appropriate threshold so that the background becomes white and the mechanism itself appears white. By applying a blur filter to the image, we obtain a fake depth which is sufficient for the Detectron2 network to identify the three links as capsules. We manually designate the instances of the segmentation map found by Detectron2 to pregenerated capsule shapes in our simulator.

Since we cannot rely on depth information to inform the 3D poses of the rigid bodies, we assume the mechanism to be a planar system. Therefore, the rigid body tracking system is constructed such that each body only has three degrees of freedom ($x$, $z$ position and yaw angle). We set up the rasterizer to produce RGB images (where the blue and red color has been manually assigned to the correct capsule shapes upfront).

\begin{figure*}
    \centering
    \begin{subfigure}[b]{0.5\textwidth}
        \centering
        \includegraphics[width=\textwidth]{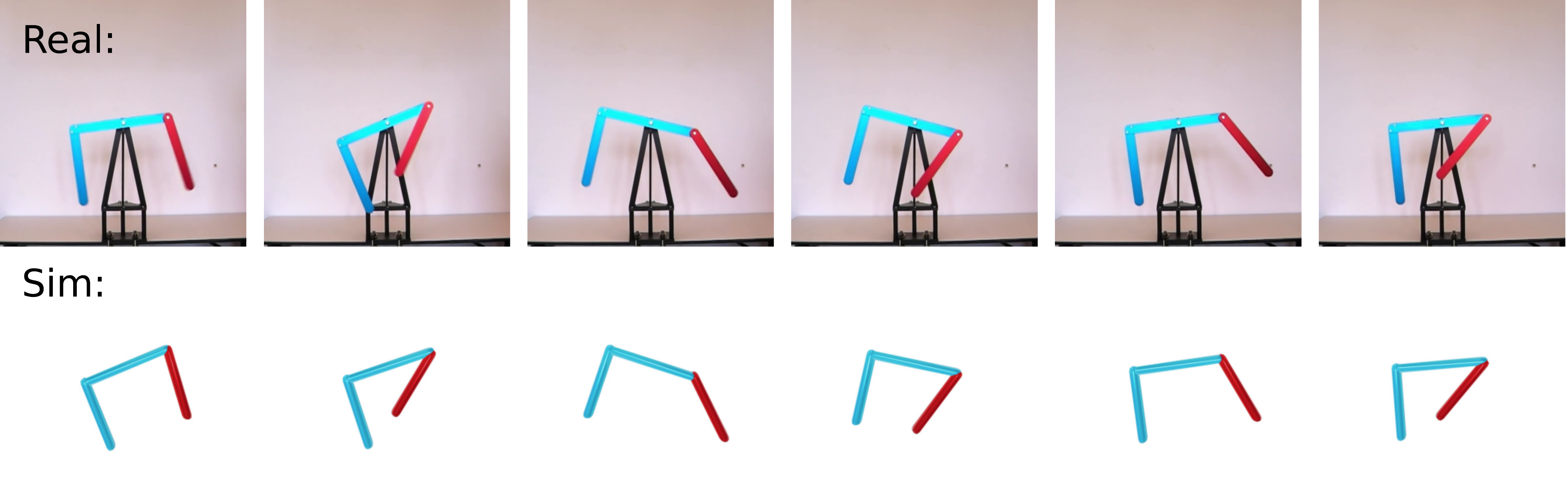}
        \caption{Simulation derived from RGB video}
        \label{fig:rott-pendulum-real}
     \end{subfigure}\hspace*{1cm}
    \begin{subfigure}[b]{0.27\textwidth}
        \centering
        \includegraphics[width=\textwidth]{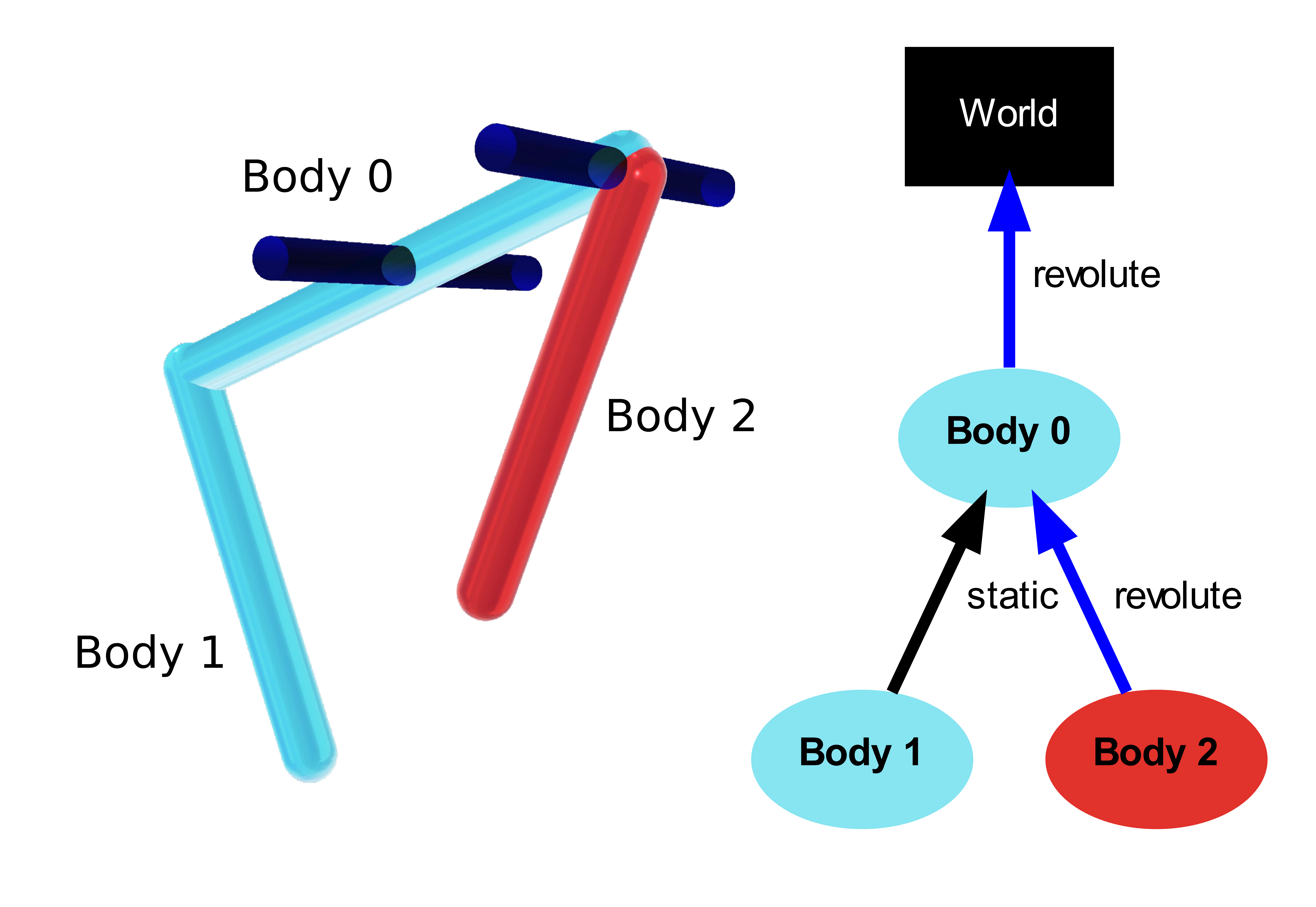}
        \caption{Inferred articulation}
        \label{fig:rott-pendulum-articulation}
     \end{subfigure}
    \caption{\textbf{Our proposed framework infers articulated rigid body dynamics simulations from video.} In this example, Rott's pendulum is identified from real RGB camera footage of the system in motion. In (a), the original video is compared to the simulated motion in the learned physics engine. In (b), the inferred joints are visualized on the left, where blue cylinders correspond to the axes of the revolute joints. The estimated kinematic tree is shown on the right.}
    \label{fig:rott-pendulum}
\end{figure*}

\begin{table}[]
    \centering
    \resizebox{\columnwidth}{!}{
    \begin{tabular}{p{0.85cm}ccccc}
        \toprule
         & \bf Adam & \bf Ours & \bf CMA-ES & \bf MCMC & \bf PhyDNet \\\midrule
        & \multicolumn{5}{c}{\bf Cartpole depth image sequence}\\
        \bf MSE & 0.0015 & 0.0008 & 0.0018 & 0.0019 & 0.1665 \\
        \bf MAE & 0.0073 & 0.0042 & 0.0089 & 0.0093 & 0.4056\\
        \bf SSIM & 0.9136 & 0.9452 & 0.8908 & 0.8893 & 0.7985 \\\midrule
        & \multicolumn{5}{c}{\bf Rott's pendulum RGB image sequence}\\
        \bf MSE & 0.0112 & 0.0134 & 0.0122 & 0.0177 & 0.0054 \\
        \bf MAE & 0.0175 & 0.0219 & 0.0187 & 0.0274 & 0.0153 \\
        \bf SSIM & 0.8929 & 0.8370 & 0.8774 & 0.8001 & 0.9412 \\
        \bottomrule
    \end{tabular}}\vspace*{0.5em}
    \caption{\textbf{Video forecasting performance.} For each of the experiments, the models were evaluated by observing the first 10 frames from the test dataset and predicting the next 150 frames.}
    \label{tab:video-forecasting}
\end{table}

\newcommand{\centered}[1]{\begin{tabular}[t]{@{}l@{}} #1 \end{tabular}}

\begin{figure}
    \centering
    \newcommand{\figwidth}{8cm}
    \resizebox{\columnwidth}{!}{
    \begin{tabular}{lc}
    \centered{Ground-truth} & \includegraphics[width=\figwidth,trim=0 0 0 2cm,clip]{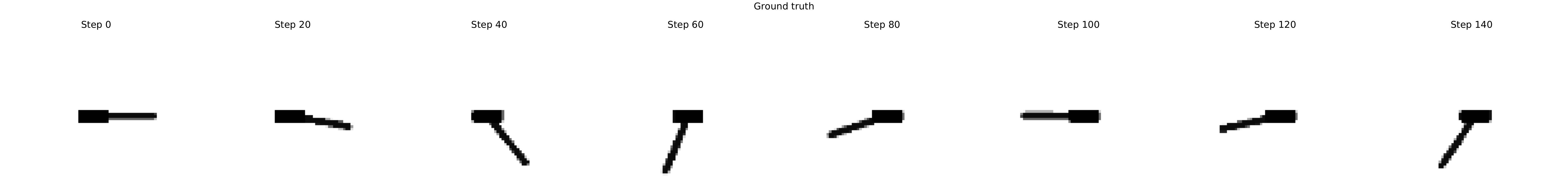} \\
    \centered{Adam} & \includegraphics[width=\figwidth,trim=0 0 0 2cm,clip]{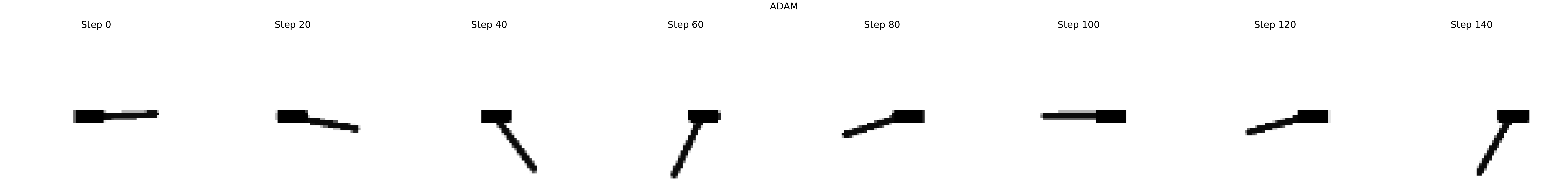} \\
    \centered{Ours} & \includegraphics[width=\figwidth,trim=0 0 0 2cm,clip]{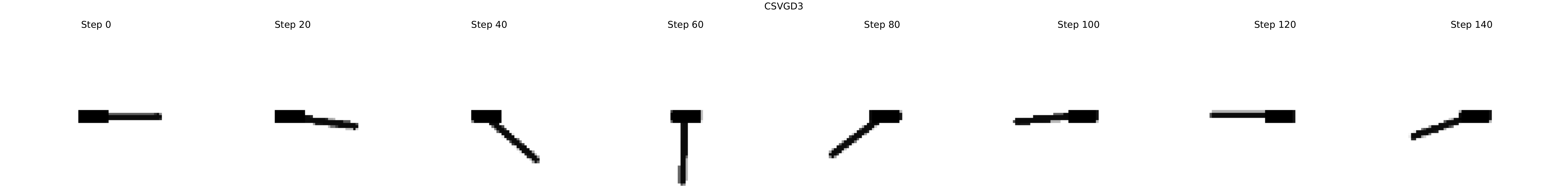} \\
    \centered{CMA-ES} & \includegraphics[width=\figwidth,trim=0 0 0 2cm,clip]{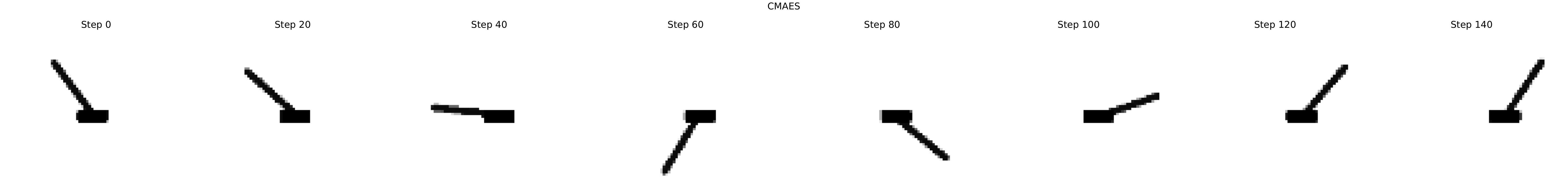} \\
    \centered{MCMC} & \includegraphics[width=\figwidth,trim=0 0 0 2cm,clip]{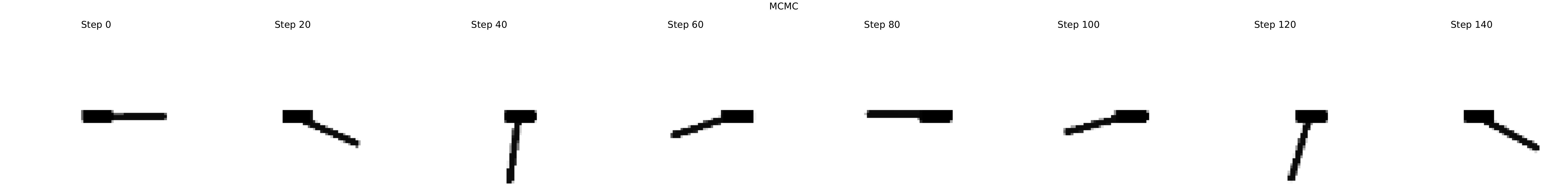} \\
    \centered{PhyDNet} & \includegraphics[width=\figwidth,trim=0 0 0 2cm,clip]{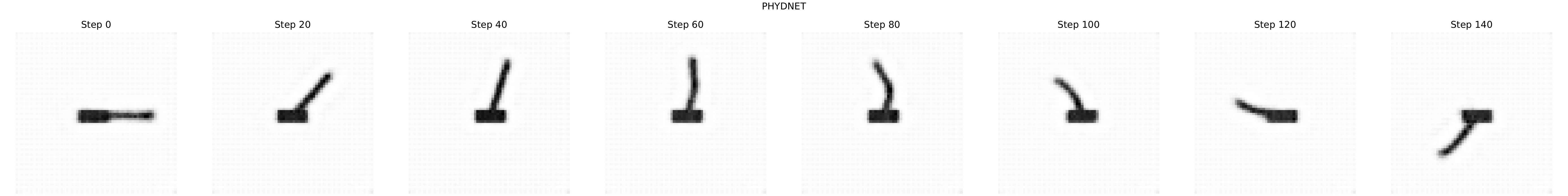}
    \end{tabular}
    }\vspace*{0.5em}
    \caption{\textbf{Video forecasting results on the test dataset of a simulated cartpole.} 150 frames must be predicted given the first 10 frames of the motion.}
    \label{fig:cartpole-forecasting}
\end{figure}

\subsection{Real articulated system}
\label{sec:exp-craftsman}

In our final experiment we reconstruct a simulation from a real-world articulated system being pushed by a robot on a table.
We build an articulated mechanism with the ``Craftsman'' toy toolkit which contains wooden levers and linkages. We command a Franka Emika Panda robot arm equipped with a cylindrical tip at its end-effector to move sideways along the $y$ direction by \SI{60}{\cm} over a duration of \SI{20}{\second} at a constant velocity.

We record RGB and depth video with a RealSense L515 LiDAR sensor. The depth images are aligned to the color images by considering the differing resolution and field-of-view of the two input modalities.
Since the depth output from the RealSense L515 LiDAR sensor does not have sufficient resolution to clearly identify the sub-centimeter parts of the Craftsman construction kit, we choose the RGB image stream as input to the instance segmentation model.
The image segmentation model is a mask R-CNN pretrained on the COCO instance segmentation dataset~\cite{msft2020coco}. We collected a dataset of 30 RGB images of three Craftsman parts (two different types of parts) in various spatial configurations for finetuning Detectron2 on our domain.

\begin{figure}
    \centering
    \newcommand{\figwidth}{\columnwidth}
    \begin{subfigure}[b]{\columnwidth}
        \centering
        \includegraphics[width=\figwidth]{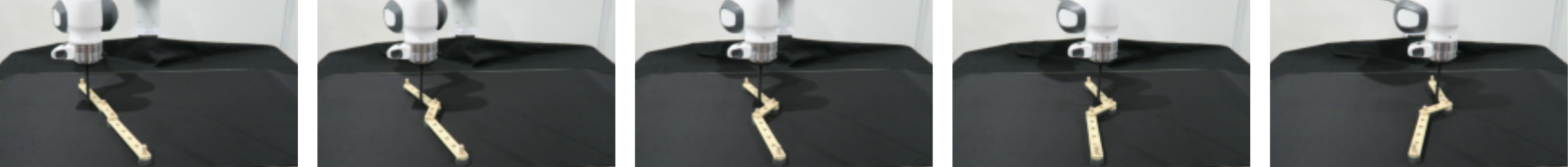}
        \caption{Real motion}
        \label{fig:craftsman-real}
     \end{subfigure}
    \begin{subfigure}[b]{\columnwidth}
        \centering
        \includegraphics[width=\figwidth]{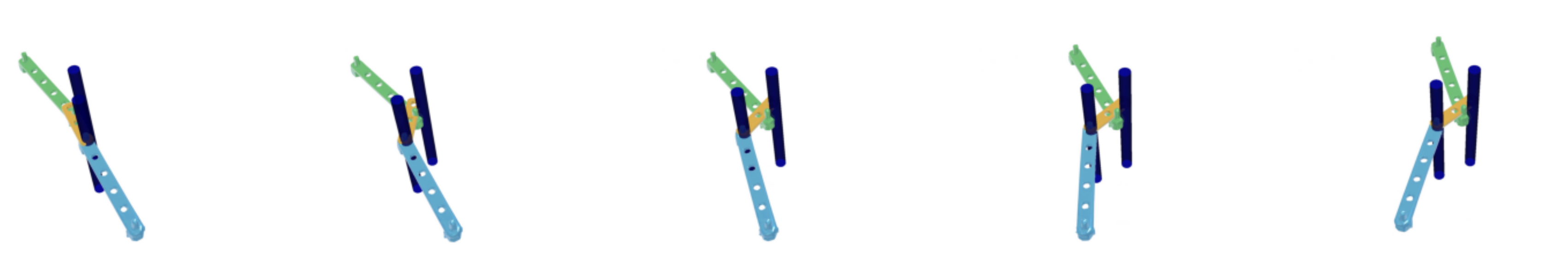}
        \caption{Inferred joints and positions}
        \label{fig:craftsman-joint-tracking}
     \end{subfigure}
    \caption{\textbf{Joint inference for the Craftsman system.} The articulation is inferred; the two revolute joints are indicated by blue cylinders in (b).}
    \label{fig:craftsman-scene-graph}
\end{figure}

Given the instance segmentation map, we extract the corresponding 3D point cloud segments for each instance (see~\autoref{fig:craftsman-segmentation}) and use iterative closest point to estimate an initial pose of each shape from the construction kit. Given this initialization, as before, the poses of the Craftsman parts are optimized through the differentiable rasterizer, where the camera has been aligned in the simulator from a perspective-n-point calibration routine. As shown in \autoref{fig:craftsman-scene-graph}, we are able to discover the two revolute joints connecting the three links of the mechanism, as well as the planar floating base (prismatic $x,y$ and revolute $z$ joints). The inferred joint positions allow the kinematic tree (shown in (b)) to move in a way that closely matches the real motion (a).

\section{Analysis}
\label{sec:results}

In this section we evaluate our approach in comparison to baselines in pose tracking, video forecasting and control.

\subsection{Rigid Pose Tracking}
\label{sec:results-rigid-tracking}

\begin{figure}
    \centering
    \includegraphics[width=0.7\columnwidth]{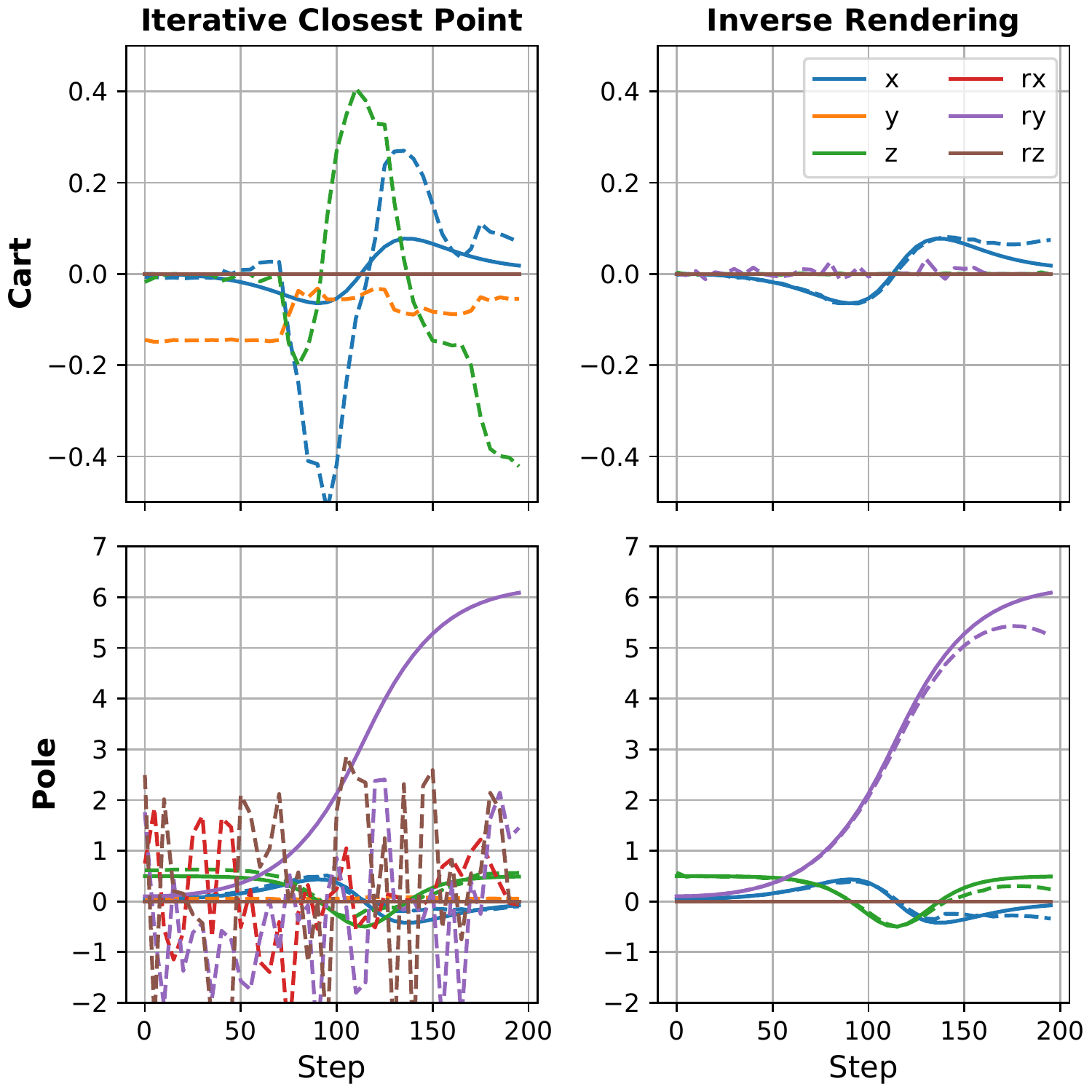}
    \caption{\textbf{Comparison.} The point-to-point iterative closest point (ICP) algorithm (left column) and our inverse rendering approach (right column) are compared on the tracking of the rigid poses of the cart (top row) and the pole (bottom row) in the simulated cartpole experiment from~\autoref{sec:exp-cartpole}. The ground-truth coordinates of the 6D poses are shown in solid lines, the corresponding estimated results are indicated by dashed lines.}
    \label{fig:cartpole-rigid-tracking}
\end{figure}

We investigate the accuracy of the rigid pose tracking phase in our approach on the simulated cartpole system which provides us access to the true 6D poses (3D positions $x,y,z$, and 3D rotations $rx, ry, rz$ in axis-angle form) of the two rigid objects (cart and pole) in the system. As shown in~\autoref{fig:cartpole-rigid-tracking}, our inverse rendering approach leveraging gradient-based optimization achieves a significantly higher accuracy compared to a classical pose tracking approach, namely point-to-point iterative closest point (ICP). For ICP, we generate point clouds from the known object meshes via Poisson disk sampling~\cite{yuksel2015poisson}, and align them to the point clouds from the depth images in the cartpole dataset (given the initial alignment at the first time step from our inverse rendering tracking approach). ICP achieves an MAE over the 6D poses of 0.755, compared to our inverse rendering approach which achieves a MAE of 0.024. The significantly higher accuracy in pose tracking is the enabler of our articulation inference phase which relies on pose estimates with few outliers and overall low noise to clearly identify the different joint types. For example, even from small changes along the $x$ axis (as in the case of the cart in the top row of \autoref{fig:cartpole-rigid-tracking}), the cart's prismatic joint can be correctly inferred due to the overall low noise level in our pose estimates.

\subsection{Video Forecasting}
\label{sec:results-video-forecasting}

In \autoref{tab:video-forecasting}, we compare the performance of our method on video forecasting against PhyDNet~\cite{DBLP:journals/corr/abs-2003-01460}. PhyDNet introduces ``PhyCells'' to disentangle physical dynamics knowledge from residual information and generates PDE-constrained future frame predictions. We extract the first 195 frames from the Rott's pendulum video and generate 175 sequences of length 20 for training, and test on frames in the later part of the video. Given the first 10 frames of the sequence as input, PhyDNet predicts the next 150 frames, which we compare against the GT frames.
The results indicate that purely vision-based model PhyDNet has an advantages where the test dataset is close to the training data, as is the case in the videos of Rott's pendulum where in both cases the pendulum calmly swings back and forth. While outperforming our simulation-based approach on all metrics, including the Structural Similarity (SSIM) index, on the cartpole test data the results clearly show an advantage of the strong inductive bias our simulation-based inference approach provides. With the accurate parameters settings (see NMAE in the top section of~\autoref{tab:cartpole-control}) our method finds, it is able to outperform all the other inference approaches on all of our video forecasting accuracy metrics.

\subsection{Model-predictive control}
\label{sec:cartpole-control}

Given the inferred simulator for the observed cartpole system, we train a controller in simulation and investigate how well it performs on the original system.
We leverage Model Predictive Path Integral (MPPI)~\cite{williams2017mppi}, a model predictive control (MPC) algorithm that has been shown to being particularly suited to control nonlinear systems.

In the swing-up task, the cartpole is resting with its pole pointing downwards. For the balancing task, the cart has a $\SI{0.1}{\meter\per\second}$ velocity applied to it, and the pole is at an angle of $20^\circ$ from its upright position at the beginning. For both tasks the goal is to bring it into an upright position while keeping the cart and pole velocities as small as possible.

The bottom two rows of \autoref{tab:cartpole-control} summarize the average rewards over a trajectory length of 200 steps (with time step \SI{0.05}{\second}) obtained by the MPPI controller running with 200 samples per step with a lookahead window length of 60 time steps. We compare the returns of the controller run on the inferred system with the most likely parameter configuration found by the different parameter inference algorithms. For reference, in the last column, we report the mean reward on the ground-truth cartpole simulation (GT).
The inferred dynamics model with the parameters found by our CSVD method allow the MPPI controller to swing up the cartpole and maintain the pole in an almost upright position without falling back down. Similarly, the balancing task succeeds after an initial phase where the pole swings downwards.

\section{Conclusions}
\label{sec:conclusion}

Our proposed pipeline allows the automatic inference of articulated rigid-body simulations from video by leveraging differentiable physics simulation and rendering. Our results on a simulated system demonstrate that we can achieve accurate trajectory predictions that benefit model-based control, while the learned parameters are physically meaningful. On a real-world coupled pendulum system, our approach predicts the correct joint topology and results in a simulation that accurately reproduces the real RGB video of the mechanism. In future work, we are planning to incorporate the learning of geometric shapes, and improve our implementation and algorithms to achieve real-time simulation inference.

\section*{Acknowledgements}
We thank Chris Denniston and David Millard for their feedback and discussion around this work, as well as Vedant Mistry for his contributions to an earlier prototype of our implementation.
We are grateful to Maria Hakuba, Hansjörg Frei and Christoph Schär for kindly permitting us to use their video of Rott's mechanism\footnote{\url{https://youtu.be/dhZxdV2naw8}. Accessed on July 28, 2022.} in this work.

\bibliographystyle{IEEEtran}
\bibliography{IEEEabrv, literature}

\end{document}

%% file: notation.tex
\usepackage[english]{babel}
\usepackage{hyperref}
\addto\captionsenglish{%
}
\addto\extrasenglish{%
}

\definecolor{darkbrown}{rgb}{0.4, 0.26, 0.13}

\newcommand{\paramdim}[0]{M}
\newcommand{\params}[0]{\theta}

\newcommand{\statevec}[0]{\mathbf{s}}

\newcommand{\observationvec}[0]{\mathbf{x}}

\newcommand{\trajectory}[0]{\mathcal{X}}
\newcommand{\trajectoryset}[0]{D_\trajectory}

%% file: root.bbl
\begin{thebibliography}{10}
\providecommand{\url}[1]{#1}
\csname url@samestyle\endcsname
\providecommand{\newblock}{\relax}
\providecommand{\bibinfo}[2]{#2}
\providecommand{\BIBentrySTDinterwordspacing}{\spaceskip=0pt\relax}
\providecommand{\BIBentryALTinterwordstretchfactor}{4}
\providecommand{\BIBentryALTinterwordspacing}{\spaceskip=\fontdimen2\font plus
\BIBentryALTinterwordstretchfactor\fontdimen3\font minus
  \fontdimen4\font\relax}
\providecommand{\BIBforeignlanguage}[2]{{%
\expandafter\ifx\csname l@#1\endcsname\relax
\typeout{** WARNING: IEEEtran.bst: No hyphenation pattern has been}%
\typeout{** loaded for the language `#1'. Using the pattern for}%
\typeout{** the default language instead.}%
\else
\language=\csname l@#1\endcsname
\fi
#2}}
\providecommand{\BIBdecl}{\relax}
\BIBdecl

\bibitem{rosinol2021kimera}
\BIBentryALTinterwordspacing
A.~Rosinol, A.~Violette, M.~Abate, N.~Hughes, Y.~Chang, J.~Shi, A.~Gupta, and
  L.~Carlone, ``Kimera: From slam to spatial perception with 3d dynamic scene
  graphs,'' \emph{The International Journal of Robotics Research}, vol.~40, no.
  12-14, pp. 1510--1546, 2021. [Online]. Available:
  \url{https://doi.org/10.1177/02783649211056674}
\BIBentrySTDinterwordspacing

\bibitem{hofer2020sim2real}
\BIBentryALTinterwordspacing
S.~H{\"{o}}fer, K.~E. Bekris, A.~Handa, J.~C.~G. Higuera, F.~Golemo,
  M.~Mozifian, C.~G. Atkeson, D.~Fox, K.~Goldberg, J.~Leonard, C.~K. Liu,
  J.~Peters, S.~Song, P.~Welinder, and M.~White, ``Perspectives on sim2real
  transfer for robotics: {A} summary of the {R:} {SS} 2020 workshop,''
  \emph{CoRR}, vol. abs/2012.03806, 2020. [Online]. Available:
  \url{https://arxiv.org/abs/2012.03806}
\BIBentrySTDinterwordspacing

\bibitem{sanchez2018graph}
A.~Sanchez-Gonzalez, N.~Heess, J.~T. Springenberg, J.~Merel, M.~Riedmiller,
  R.~Hadsell, and P.~Battaglia, ``Graph networks as learnable physics engines
  for inference and control,'' \emph{arXiv preprint arXiv:1806.01242}, 2018.

\bibitem{sanchez2020learning}
A.~Sanchez-Gonzalez, J.~Godwin, T.~Pfaff, R.~Ying, J.~Leskovec, and P.~W.
  Battaglia, ``Learning to simulate complex physics with graph networks,''
  \emph{arXiv preprint arXiv:2002.09405}, 2020.

\bibitem{murthy2021gradsim}
J.~K. Murthy, M.~Macklin, F.~Golemo, V.~Voleti, L.~Petrini, M.~Weiss,
  B.~Considine, J.~Parent-L{\'e}vesque, K.~Xie, K.~Erleben, L.~Paull,
  F.~Shkurti, D.~Nowrouzezahrai, and S.~Fidler, ``grad{S}im: Differentiable
  simulation for system identification and visuomotor control,'' in
  \emph{International Conference on Learning Representations}, 2021.

\bibitem{ding2021vrdp}
M.~Ding, Z.~Chen, T.~Du, P.~Luo, J.~B. Tenenbaum, and C.~Gan, ``Dynamic visual
  reasoning by learning differentiable physics models from video and
  language,'' in \emph{Advances In Neural Information Processing Systems},
  2021.

\bibitem{wu2015galileo}
J.~Wu, I.~Yildirim, J.~J. Lim, W.~T. Freeman, and J.~B. Tenenbaum, ``Galileo:
  Perceiving physical object properties by integrating a physics engine with
  deep learning,'' in \emph{Advances in Neural Information Processing Systems},
  2015, pp. 127--135.

\bibitem{jain20screwnet}
A.~Jain, R.~Lioutikov, C.~Chuck, and S.~Niekum, ``Screwnet:
  Category-independent articulation model estimation from depth images using
  screw theory,'' in \emph{arXiv preprint}, 2020.

\bibitem{hausman2015articulation}
\BIBentryALTinterwordspacing
K.~Hausman, S.~Niekum, S.~Osentoski, and G.~S. Sukhatme, ``Active articulation
  model estimation through interactive perception,'' in \emph{2015 IEEE
  International Conference on Robotics and Automation (ICRA)}.\hskip 1em plus
  0.5em minus 0.4em\relax IEEE, May 2015, p. 3305–3312. [Online]. Available:
  \url{http://ieeexplore.ieee.org/document/7139655/}
\BIBentrySTDinterwordspacing

\bibitem{eppner2018physics}
C.~Eppner, R.~Mart\'{\i}n-Mart\'{\i}n, and O.~Brock, ``Physics-based selection
  of informative actions for interactive perception,'' in \emph{Proceedings of
  the IEEE International Conference on Robotics and Automation (ICRA)}, 2018.

\bibitem{martin2014articulation}
\BIBentryALTinterwordspacing
R.~Martin~Martin and O.~Brock, ``Online interactive perception of articulated
  objects with multi-level recursive estimation based on task-specific
  priors,'' in \emph{2014 IEEE/RSJ International Conference on Intelligent
  Robots and Systems}.\hskip 1em plus 0.5em minus 0.4em\relax IEEE, Sep 2014,
  p. 2494–2501. [Online]. Available:
  \url{https://ieeexplore.ieee.org/document/6942902/}
\BIBentrySTDinterwordspacing

\bibitem{sturm2011articulation}
J.~Sturm, C.~Stachniss, and W.~Burgard, ``A probabilistic framework for
  learning kinematic models of articulated objects,'' \emph{J. Artif. Int.
  Res.}, vol.~41, no.~2, p. 477–526, may 2011.

\bibitem{niekum2015online}
S.~Niekum, S.~Osentoski, C.~G. Atkeson, and A.~G. Barto, ``Online bayesian
  changepoint detection for articulated motion models,'' in \emph{2015 IEEE
  International Conference on Robotics and Automation (ICRA)}.\hskip 1em plus
  0.5em minus 0.4em\relax IEEE, 2015, pp. 1468--1475.

\bibitem{liu2019articulation}
Y.~Liu, F.~Zha, L.~Sun, J.~Li, M.~Li, and X.~Wang, ``Learning articulated
  constraints from a one-shot demonstration for robot manipulation planning,''
  \emph{IEEE Access}, vol.~7, p. 172584–172596, 2019.

\bibitem{jain2021distributional}
\BIBentryALTinterwordspacing
A.~Jain, S.~Giguere, R.~Lioutikov, and S.~Niekum, ``Distributional depth-based
  estimation of object articulation models,'' in \emph{5th Annual Conference on
  Robot Learning}, 2021. [Online]. Available:
  \url{https://openreview.net/forum?id=H1-uwiTbY9z}
\BIBentrySTDinterwordspacing

\bibitem{mu2021asdf}
J.~Mu, W.~Qiu, A.~Kortylewski, A.~Yuille, N.~Vasconcelos, and X.~Wang, ``A-sdf:
  Learning disentangled signed distance functions for articulated shape
  representation,'' \emph{arXiv preprint arXiv: 2104.07645}, 2021.

\bibitem{noguchi2021watch}
A.~Noguchi, U.~Iqbal, J.~Tremblay, T.~Harada, and O.~Gallo, ``Watch it move:
  Unsupervised discovery of 3d joints for re-posing of articulated objects,''
  2021.

\bibitem{battaglia2016interaction}
P.~Battaglia, R.~Pascanu, M.~Lai, D.~J. Rezende \emph{et~al.}, ``Interaction
  networks for learning about objects, relations and physics,'' in
  \emph{Advances in Neural Information Processing Systems}, 2016, pp.
  4502--4510.

\bibitem{xu2019physics}
Z.~Xu, J.~Wu, A.~Zeng, J.~B. Tenenbaum, and S.~Song, ``{DensePhysNet}: Learning
  dense physical object representations via multi-step dynamic interactions,''
  \emph{Robotics: Science and Systems}, Jun. 2019.

\bibitem{sanchezgonzalez2020learning}
A.~Sanchez-Gonzalez, J.~Godwin, T.~Pfaff, R.~Ying, J.~Leskovec, and P.~W.
  Battaglia, ``Learning to simulate complex physics with graph networks,''
  2020.

\bibitem{raissi2019physics}
M.~Raissi, P.~Perdikaris, and G.~E. Karniadakis, ``Physics-informed neural
  networks: A deep learning framework for solving forward and inverse problems
  involving nonlinear partial differential equations,'' \emph{Journal of
  Computational Physics}, vol. 378, pp. 686--707, 2019.

\bibitem{lutter2019delan}
M.~Lutter, C.~Ritter, and J.~Peters, ``Deep {L}agrangian networks: Using
  physics as model prior for deep learning,'' \emph{arXiv preprint
  arXiv:1907.04490}, 2019.

\bibitem{sutanto20rnea}
\BIBentryALTinterwordspacing
G.~Sutanto, A.~Wang, Y.~Lin, M.~Mukadam, G.~Sukhatme, A.~Rai, and F.~Meier,
  ``Encoding physical constraints in differentiable {N}ewton-{E}uler
  algorithm,'' ser. Proceedings of Machine Learning Research, A.~M. Bayen,
  A.~Jadbabaie, G.~Pappas, P.~A. Parrilo, B.~Recht, C.~Tomlin, and
  M.~Zeilinger, Eds., vol. 120.\hskip 1em plus 0.5em minus 0.4em\relax The
  Cloud: PMLR, 10--11 Jun 2020, pp. 804--813. [Online]. Available:
  \url{http://proceedings.mlr.press/v120/sutanto20a.html}
\BIBentrySTDinterwordspacing

\bibitem{peres2018lcp}
F.~de~Avila Belbute-Peres, K.~Smith, K.~Allen, J.~Tenenbaum, and J.~Z. Kolter,
  ``End-to-end differentiable physics for learning and control,'' in
  \emph{Advances in Neural Information Processing Systems 31}, 2018, pp.
  7178--7189.

\bibitem{qiao2020scalable}
Y.-L. Qiao, J.~Liang, V.~Koltun, and M.~C. Lin, ``Scalable differentiable
  physics for learning and control,'' in \emph{ICML}, 2020.

\bibitem{geilinger2020add}
M.~Geilinger, D.~Hahn, J.~Zehnder, M.~Bächer, B.~Thomaszewski, and S.~Coros,
  ``Add: Analytically differentiable dynamics for multi-body systems with
  frictional contact,'' in \emph{arXiv}, 2020.

\bibitem{heiden2020lidar}
E.~Heiden, Z.~Liu, R.~K. Ramachandran, and G.~S. Sukhatme, ``Physics-based
  simulation of continuous-wave {LIDAR} for localization, calibration and
  tracking,'' in \emph{International Conference on Robotics and Automation
  (ICRA)}.\hskip 1em plus 0.5em minus 0.4em\relax IEEE, 2020.

\bibitem{wu2019detectron2}
Y.~Wu, A.~Kirillov, F.~Massa, W.-Y. Lo, and R.~Girshick, ``Detectron2,''
  \url{https://github.com/facebookresearch/detectron2}, 2019.

\bibitem{laine2020nvdiffrast}
S.~Laine, J.~Hellsten, T.~Karras, Y.~Seol, J.~Lehtinen, and T.~Aila, ``Modular
  primitives for high-performance differentiable rendering,'' \emph{ACM
  Transactions on Graphics}, vol.~39, no.~6, 2020.

\bibitem{fischler1981ransac}
\BIBentryALTinterwordspacing
M.~A. Fischler and R.~C. Bolles, ``Random sample consensus: A paradigm for
  model fitting with applications to image analysis and automated
  cartography,'' \emph{Commun. ACM}, vol.~24, no.~6, p. 381–395, jun 1981.
  [Online]. Available: \url{https://doi.org/10.1145/358669.358692}
\BIBentrySTDinterwordspacing

\bibitem{heiden2021neuralsim}
\BIBentryALTinterwordspacing
E.~Heiden, D.~Millard, E.~Coumans, Y.~Sheng, and G.~S. Sukhatme, ``Neural{S}im:
  Augmenting differentiable simulators with neural networks,'' in
  \emph{Proceedings of the IEEE International Conference on Robotics and
  Automation (ICRA)}, 2021. [Online]. Available:
  \url{https://github.com/google-research/tiny-differentiable-simulator}
\BIBentrySTDinterwordspacing

\bibitem{featherstone2007rbda}
R.~Featherstone, \emph{Rigid Body Dynamics Algorithms}.\hskip 1em plus 0.5em
  minus 0.4em\relax Berlin, Heidelberg: Springer-Verlag, 2007.

\bibitem{heiden2022pds}
\BIBentryALTinterwordspacing
E.~Heiden, C.~E. Denniston, D.~Millard, F.~Ramos, and G.~S. Sukhatme,
  ``Probabilistic inference of simulation parameters via parallel
  differentiable simulation,'' \emph{CoRR}, vol. abs/2109.08815, 2021.
  [Online]. Available: \url{https://arxiv.org/abs/2109.08815}
\BIBentrySTDinterwordspacing

\bibitem{coumans2013bullet}
E.~Coumans \emph{et~al.}, ``Bullet physics library,'' \emph{Open source:
  bulletphysics.org}, vol.~15, no.~49, p.~5, 2013.

\bibitem{hansen2016cma}
N.~Hansen, ``The cma evolution strategy: A tutorial,'' 2016.

\bibitem{foreman2013emcee}
\BIBentryALTinterwordspacing
D.~Foreman-Mackey, D.~W. Hogg, D.~Lang, and J.~Goodman, ``emcee: The {MCMC}
  hammer,'' \emph{Publications of the Astronomical Society of the Pacific},
  vol. 125, no. 925, p. 306–312, Mar 2013. [Online]. Available:
  \url{http://dx.doi.org/10.1086/670067}
\BIBentrySTDinterwordspacing

\bibitem{rott1970pendulum}
N.~Rott, ``A multiple pendulum for the demonstration of non-linear coupling,''
  \emph{Zeitschrift f{\"u}r angewandte Mathematik und Physik ZAMP}, vol.~21,
  no.~4, pp. 570--582, 1970.

\bibitem{msft2020coco}
T.-Y. Lin, M.~Maire, S.~Belongie, J.~Hays, P.~Perona, D.~Ramanan, P.~Dollár,
  and C.~L. Zitnick, ``Microsoft coco: Common objects in context,'' in
  \emph{Computer Vision – ECCV 2014}, D.~Fleet, T.~Pajdla, B.~Schiele, and
  T.~Tuytelaars, Eds.\hskip 1em plus 0.5em minus 0.4em\relax Springer
  International Publishing, 2014, p. 740–755.

\bibitem{yuksel2015poisson}
C.~Yuksel, ``Sample elimination for generating poisson disk sample sets,''
  \emph{Computer Graphics Forum}, vol.~34, no.~2, p. 25–32, 2015.

\bibitem{DBLP:journals/corr/abs-2003-01460}
\BIBentryALTinterwordspacing
V.~L. Guen and N.~Thome, ``Disentangling physical dynamics from unknown factors
  for unsupervised video prediction,'' \emph{CoRR}, vol. abs/2003.01460, 2020.
  [Online]. Available: \url{https://arxiv.org/abs/2003.01460}
\BIBentrySTDinterwordspacing

\bibitem{williams2017mppi}
G.~Williams, N.~Wagener, B.~Goldfain, P.~Drews, J.~M. Rehg, B.~Boots, and E.~A.
  Theodorou, ``Information theoretic mpc for model-based reinforcement
  learning,'' in \emph{2017 IEEE International Conference on Robotics and
  Automation (ICRA)}, 2017, pp. 1714--1721.

\end{thebibliography}
